\documentclass[cameraready]{Interspeech}
\title{Adaptive Speech-to-Spike Encoding for Spiking Neural Networks}

\author[affiliation={1}, orcid=0009-0006-4918-7920]{Taharim Rahman}{Anon}
\author[affiliation={1}, orcid=0009-0002-3740-3818]{Jakaria Islam}{Emon}

\address{
$^{1}$PI LLC, Sapporo, Hokkaido, Japan
}

\email{tahrim.anon21@gmail.com, emon@pisapporo.com}


\usepackage{comment}
\usepackage{graphicx}
\usepackage{bbm}
\graphicspath{{figures/}}
\usepackage{needspace}

\newcommand{\vect}[1]{\bm{#1}}
\newcommand{\R}{\mathbb{R}}

\newcommand{\Hstep}{H} % Heaviside

\newcommand\blfootnote[1]{%
  \begingroup
  \renewcommand\thefootnote{}\footnote{#1}%
  \addtocounter{footnote}{-1}%
  \endgroup
}
\begin{document}

\maketitle
\blfootnote{This paper was accepted at Interspeech 2026. This version is a preprint.}
%==================================================================================
\begin{abstract}
The mismatch between continuous acoustic signals and discrete event-driven processing remains a fundamental bottleneck for neuromorphic speech processing. Current systems typically rely on fixed spike encoders, forcing downstream Spiking Neural Networks (SNNs) to compensate for non-adaptive input representations. To address this, we present a learnable residual speech-to-spike encoder jointly trained end-to-end with a Recurrent Leaky Integrate-and-Fire (R-LIF) backbone. 
We validate this approach on the Google Speech Commands v2 (GSC-v2) benchmark, achieving up to \textbf{94.97\%} accuracy. Notably, the learned encoder remains highly parameter-efficient with a compact 35k-parameter variant that reaches \textbf{89.8\%}, matching or exceeding prior baselines that require an order of magnitude more parameters. 
Our encoder-focused analysis, including linear probing and gradient-residual inspection, indicates that the encoder does not target faithful signal reconstruction but instead learns task-aligned spike representations that enhance class separability. 
Finally, we benchmark bio-inspired, hardware-friendly credit assignment by comparing Direct Feedback Alignment (DFA) with surrogate-gradient BPTT under identical architectures and training conditions. We find that DFA reaches \textbf{91.5\%} accuracy, quantifying the performance trade-off of bio-inspired learning rules for modern neuromorphic audio.

\end{abstract}
%==================================================================================
\section{Introduction}
\label{sec:intro}

Neuromorphic computing offers a compelling paradigm for processing temporal signals at the extreme edge, promising high energy efficiency through sparse, event-driven processing.
However, mapping continuous-time auditory signals to discrete Spiking Neural Networks (SNNs) remains a fundamental challenge. Unlike the visual domain, where dynamic vision sensors (DVS) \cite{lichtsteiner2008dvs} provide a native asynchronous event stream, audio acquisition typically relies on standard microphones that output dense, high-bandwidth waveforms. In practice, fixed step-forward speech-to-spike encoders rely on static, manually chosen thresholds. Consequently, even advanced SNN architectures with convolutional~\cite{ylmaz20_interspeech,wang24p_interspeech} or recurrent~\cite{bittar22_frontiers_sg_baseline} backbones remain constrained by this fixed, heuristic encoder \cite{article}.
This fixed structure forces the network to compensate for suboptimal input representations, often requiring larger models to extract discriminative features.
 
Parallel to these architectural constraints is the challenge of learning with local, hardware-compatible update rules. While Backpropagation Through Time (BPTT) with surrogate-gradient remains the gold standard for accuracy ~\cite{zenke21_neuralcomputation_surrogates}, it relies on computationally expensive error propagation and symmetric weights that are difficult to implement on neuromorphic hardware \cite{osti_2476747}. Bio-inspired alternatives such as Direct Feedback Alignment (DFA)~\cite{nokland16_neurips_dfa,lillicrap16_natcomm_feedback_alignment} avoid the weight transport problem and enable parallel layer updates, including recent extensions operating directly on spike trains \cite{Lee2020SpikeTrainST-DFA}. Yet the performance trade-off between these feedback mechanisms and surrogate BPTT on neuromorphic speech tasks remains underexplored.
 
Motivated by these limitations, we introduce a fully differentiable residual speech-to-spike encoding mechanism that replaces fixed thresholds with learnable parameters. We validate this approach on the Google Speech Commands v2 (GSC-v2) dataset \cite{warden2018speechcommandsdatasetlimitedvocabulary}, a stringent benchmark where traditional deep learning  baselines set a high bar~\cite{gong21b_interspeech,berg21_interspeech}. We summarize our contributions as follows: 

\begin{enumerate}[leftmargin=*]
  \item We introduce a learnable residual speech-to-spike encoder, jointly optimized with a recurrent LIF (R-LIF) backbone. This yields up to 94.97\% accuracy with an average encoder spike rate of 6.56\%, enabling competitive performance at small model sizes down to 35k parameters.
  \item We provide an encoder-side analysis using temporal probes and gradient statistics and demonstrate that the learnable encoder does not aim to faithfully reconstruct the input log-mel spectrogram; instead, it constructs a task-aligned spike representation that significantly enhances the linear separability of classes compared to fixed baselines.
  \item We benchmark credit-assignment mechanisms by comparing surrogate-gradient BPTT with DFA. Under this comparison, DFA reaches 91.5\% accuracy (vs. 94.97 for BPTT) for spiking keyword spotting  on the GSC dataset, clarifying both the potential and current limitations of bio-inspired learning rules.
\end{enumerate}

The rest of the paper is organized as follows: Section~\ref{sec:method} details the learnable residual encoder and R-LIF backbone. Section~\ref{sec:exp} outlines the experimental setup and training protocols. Section~\ref{sec:results} presents performance benchmarks, providing an interpretability analysis of the proposed encoder and evaluating learning rule trade-offs. Finally, Section~\ref{sec:conclusion} concludes with future directions.
%==================================================================================

\section{Methods}
\label{sec:method}
\begin{figure}[t]
  \centering
  \includegraphics[width=\columnwidth]{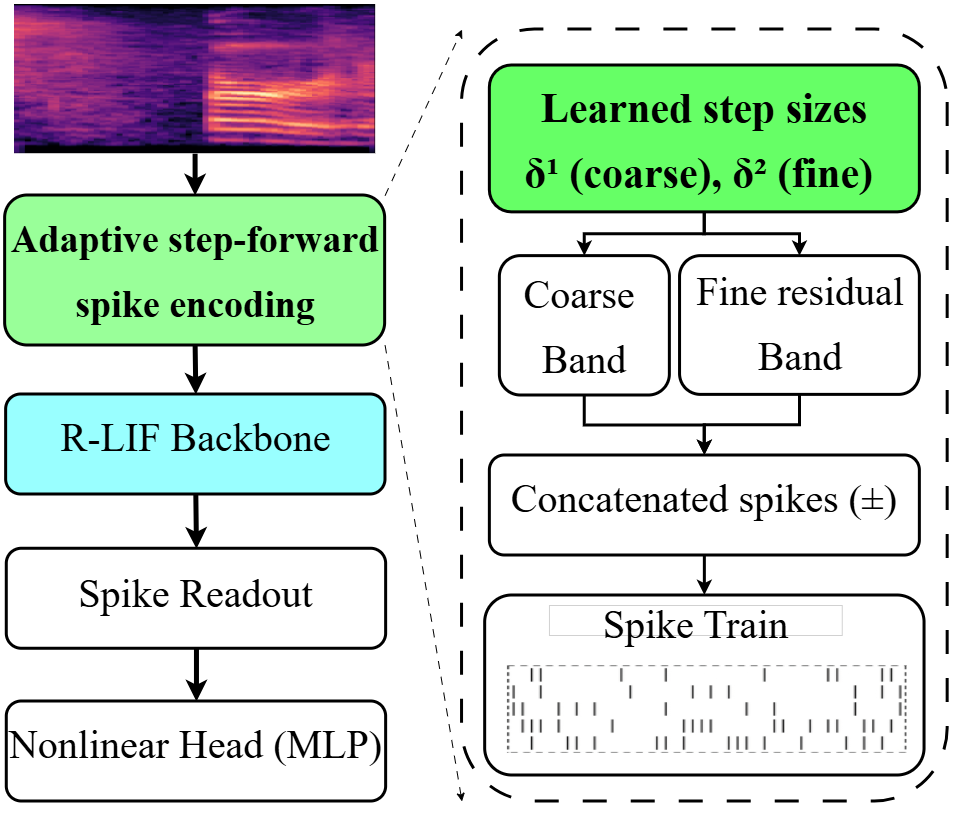}
  \caption{Overview of the proposed learnable step-forward speech-to-spike (S2S) encoder with a spiking backbone. The encoder converts log-mel features into signed spike trains using learned step sizes $\delta^{1}$ (coarse) and $\delta^{2}$ (fine residual). The resulting event streams are processed by an R-LIF backbone, followed by spike readout and a lightweight MLP head.}
  \label{fig:architecture}
\end{figure}

As illustrated in Fig.~\ref{fig:architecture}, the proposed architecture integrates a differentiable speech-to-spike front end with a recurrent spiking classifier. The pipeline processes an input log-mel spectrogram $\vect{X}\in\R^{C\times T}$, where $x_{c,t}$ denotes the log-mel magnitude at band $c$ and time $t$. At the encoder stage, the  dense input is converted into a sparse binary spike tensor $\vect{S}$ by the learnable step-forward encoder (Sec.~\ref{ssec:encoder}). The encoder emits positive and negative spikes for both coarse and fine streams, producing a 4C-channel event representation. 
Event streams are processed by a multi-layer Recurrent LIF backbone (Sec.~\ref{ssec:rlif}) to capture temporal dependencies and a lightweight nonlinear readout head maps the aggregated spike activity to class probability. 
The full pipeline is differentiable, allowing the encoder parameters to be optimized jointly with the backbone weights.

\subsection{Learnable Step-Forward Encoding}
\label{ssec:encoder}
The encoder transforms the sequence of log-mel feature values $x_{c,t}$ for each frequency band into spike trains. For clarity, we omit the frequency-band index $c$ in the equations below. The encoder maintains a coarse trace $\hat{x}_t$ with a larger step size $\delta^{(1)}$ and a fine residual trace $\hat{r}_t$ with a smaller step size $\delta^{(2)}$. 
For each frequency band, these traces are initialized to zero and are not learned parameters. The step-size parameters $\rho$ and $\kappa$ are global scalar variables shared across all frequency bands.

\vspace{0.5mm}
\noindent\textbf{Phase I: Coarse Encoding.}
We first calculate the tracking error relative to the coarse trace: $e^{(1)}_t = x_t - \hat{x}_{t-1}$.
Spikes are emitted if the error magnitude exceeds the coarse step:
\begin{align}
  s^{(1,+)}_t &= \Hstep\bigl(e^{(1)}_t-\delta^{(1)}\bigr), \\
  s^{(1,-)}_t &= \Hstep\bigl(-e^{(1)}_t-\delta^{(1)}\bigr).
\end{align}
where $\Hstep(\cdot)$ is the Heaviside step function. The coarse trace is updated via a Step-Forward rule:
\begin{equation}
  \hat{x}_t = \hat{x}_{t-1} + \delta^{(1)} \bigl(s^{(1,+)}_t - s^{(1,-)}_t\bigr).
\end{equation}

\vspace{0.5mm}
\noindent\textbf{Phase II: Fine Residual Encoding.}
After applying coarse correction, we compute the time-step residual as, $r_t = x_t - \hat{x}_t$. The fine error is then $e^{(2)}_t = r_t - \hat{r}_{t-1}$ and encoded with the finer step $\delta^{(2)}$:
\Needspace{5\baselineskip}
\begin{align}
  s^{(2,+)}_t &= \Hstep\bigl(e^{(2)}_t-\delta^{(2)}\bigr), \\
  s^{(2,-)}_t &= \Hstep\bigl(-e^{(2)}_t-\delta^{(2)}\bigr).
\end{align}
The fine trace is updated similarly:
\begin{equation}
  \hat{r}_t = \hat{r}_{t-1} + \delta^{(2)} \bigl(s^{(2,+)}_t - s^{(2,-)}_t\bigr).
\end{equation}
Final spike output is the concatenation of these four streams: $\vect{s}_t = \bigl[s^{(1,\pm)}_t, s^{(2,\pm)}_t\bigr]$ resulting in 4C channels per time step.

\vspace{0.5mm}
\noindent\textbf{Optimization and Hierarchy.}
To ensure a valid  hierarchy between step sizes ($\delta^{(2)} < \delta^{(1)}$), we parameterize the step sizes using trainable scalar variables $\rho, \kappa \in \mathbb{R}$ that are shared across all frequency bands:
\begin{equation}
  \delta^{(1)} = \operatorname{softplus}(\rho) + \epsilon,\qquad
  \delta^{(2)} = \delta^{(1)} \cdot \sigma(\kappa),
\end{equation}
where $\sigma(\cdot)$ is the logistic sigmoid and $\epsilon = 10^{-4}$ is a small constant ensuring strictly
positive step sizes.   
Since $\Hstep(\cdot)$ is non-differentiable, we use a surrogate gradient for backpropagation
\cite{8891809}. Specifically, we employ a straight-through estimator with a sigmoid backward shape in the encoder.

\subsection{Recurrent LIF Backbone and Nonlinear Readout}
\label{ssec:rlif}

The spiking backbone consists of a stack of R-LIF layers.
At layer $\ell$, let $s^{\ell}_t\in\{0,1\}^{d_\ell}$ be the spike output and $u^{\ell}_t\in\R^{d_\ell}$ the membrane potential.
Given the input $s^{\ell-1}_t$, the dynamics are governed by:
\begin{align}
  u^{\ell}_t &= \beta^\ell \odot v^{\ell}_{t-1} + W^\ell s^{\ell-1}_t + W^\ell_{\mathrm{rec}} s^{\ell}_{t-1} + b^\ell, \\
  s^{\ell}_t &= \Hstep\bigl(u^{\ell}_t - \theta^\ell\bigr), \\
  v^{\ell}_t &= u^{\ell}_t - \theta^\ell \odot s^{\ell}_t,
\end{align}
where $\beta^{\ell}$ is a learnable leak, $\theta^{\ell}$ is a learnable threshold, and $v^{\ell}_{t}$ denotes the post-reset membrane state.
Eq. (10) implements a subtractive (soft) reset. For the backbone, we differentiate spikes using the fast-sigmoid surrogate~\cite{zenke21_neuralcomputation_surrogates}.
We summarize the final-layer spike train via time averaging,
\begin{equation}
  \bar{s} = \frac{1}{T}\sum_{t=1}^{T} s^L_t,
\end{equation}
and map $\bar{s}$ to class logits using a lightweight two-layer MLP:
\begin{equation}
  \mathrm{logits} = W_2\,\phi\bigl(\mathrm{Dropout}(W_1 \bar{s} + b_1)\bigr) + b_2,
\end{equation}
where $\phi(\cdot)$ is ReLU.
As this head runs once per utterance, it adds minimal compute while improving separability compared to a purely linear readout.

\subsection{Training Objective and Learning Algorithms}
\label{sec:learning}

The training objective combines cross-entropy with an activity penalty to induce sparsity:
\begin{equation}
\mathcal{L}
=
\mathcal{L}_{\mathrm{CE}}
+
\lambda_{\mathrm{spk}}
\sum_{\ell}
\frac{1}{T d_{\ell}}
\sum_{t=1}^{T}
\left\lVert s_{t}^{\ell} \right\rVert_{1}.
\end{equation}
where $s_t^\ell \in \{0,1\}^{d_\ell}$ denotes the spike vector at layer $\ell$ and time $t$, and $\lVert s_t^\ell \rVert_1$ counts the number of spikes at that layer and time step. Here, $d_\ell$ is the number of neurons in layer $\ell$, and $T$ is the number of discrete simulation steps. We compare two credit assignment strategies to optimize this objective:

\vspace{0.5mm}
\noindent\textbf{Surrogate-Gradient BPTT.} 
we trained the network with BPTT \cite{zenke21_neuralcomputation_surrogates}, propagating gradients through the unrolled temporal graph. The discrete spiking thresholds are handled using the surrogate derivative of $\Hstep(\cdot)$.

\vspace{0.5mm}
\noindent\textbf{Direct Feedback Alignment (DFA).}
To assess hardware-efficient training, we alternatively employ DFA~\cite{nokland16_neurips_dfa}. This decouples the backward pass by projecting a global error signal $\mathbf{e}$ to hidden layers via fixed random matrices $B^\ell$. The global error $\mathbf{e} = \partial \mathcal{L} / \partial \mathbf{z}$ denotes the gradient of the loss with respect to the output logits $\mathbf{z}$, and has dimension equal to the number of classes. The local error signal is:
\begin{equation}
  \delta^\ell_t = (B^\ell \mathbf{e}) \odot \tilde{\Hstep}'(u^\ell_t - \theta^\ell).
\end{equation}
This avoids the symmetric weight transport required by BPTT while still using a local surrogate gradient for the spiking nonlinearity.

\section{Experimental Setup}
\label{sec:exp}

We evaluate on the Google Speech Commands v2 (GSC-v2) dataset using the standard 35-class protocol~\cite{warden2018speechcommandsdatasetlimitedvocabulary}. Audio is sampled at 16\,kHz. We extract 80-bin log-mel spectrograms using a 25\,ms analysis window and a 10\,ms hop, and apply log compression ($\log(1+x)$) to the mel power spectrum before passing features to the spike encoder.

The encoder maps 80 input bins to 320 spike channels (coarse± and fine±). We evaluate three backbone scales with identical R-LIF dynamics: \textit{Large} ($\sim$1.8M parameters), \textit{Small} ($\sim$0.7M parameters), and a highly constrained \textit{Tiny} model ($\sim$35k parameters). We compare the proposed \emph{learnable} encoder against two baselines: (i) the same R-LIF backbone driven by a fixed Step-Forward encoder employing non-learned thresholds, and (ii) state-of-the-art spiking KWS systems from prior work.

All models are trained with AdamW and cosine learning-rate decay\cite{loshchilov2018decoupled}. We apply standard regularization, including dropout and label smoothing, together with the spike-rate penalty described in Sec.~\ref{sec:method}. To maximize performance under tight parameter constraints, we use knowledge distillation, the compact student model minimizes the KL divergence between its temperature-scaled probabilities and those of a pre-trained teacher model, in addition to the standard cross-entropy loss.

%==================================================================================
\section{Results and Analysis}
\label{sec:results}

\subsection{Efficacy of Learnable Encoding}
\label{ssec:enc_ablation}

We first isolate the contribution of the proposed encoder by comparing it against a fixed Step-Forward baseline under an identical R-LIF backbone. As shown in Table~\ref{tab:encoder_comparison}, proposed learnable encoder yields a substantial gain in test accuracy, improving performance from 90.70\% to 94.97\%.
This improvement is accompanied by a substantial reduction in input event
activity (2982 $\rightarrow$ 2119 spikes per utterance). Consistent with this
trend, the measured input sparsity increases from 90.7\% to 93.4\%, indicating
that the learned thresholds reduce unnecessary events while preserving
task-relevant information.
Together, these results indicate that end-to-end optimization can regulate the spike budget more effectively than static, non-adaptive threshold settings.

\begin{table}[ht]
  \centering
  \caption{ Comparison of fixed vs.\ learnable encoding under an identical R-LIF backbone. Fixed Step Size refers to the hand tuned Step-Forward quantization step $\Delta$ (threshold) used in the baseline encoder.}
  \label{tab:encoder_comparison}
  \small
  \setlength{\tabcolsep}{3pt}
  \begin{tabular}{@{}lcc@{}}
    \toprule
    \textbf{Metric} & \textbf{\shortstack{Step-Forward\\(Fixed Step Size)}} &
    \textbf{\shortstack{Learnable \\S2S (Ours)}} \\ 
    \midrule
    Test Acc. (\%)           & 90.70 & \textbf{94.97} \\
    Spikes / utterance       & 2982  & \textbf{2119}  \\
    Input Sparsity (\%)      & 90.7  & \textbf{93.4}  \\
    
    \bottomrule
  \end{tabular}
\end{table}

\subsection{Comparison with State-of-the-Art}
\label{ssec:scaling}

Table~\ref{tab:kws_models} benchmarks our system against recent spiking KWS baselines on GSC-v2.
In the Large configuration, our model reaches 94.97\% accuracy, outperforming prior spiking pipelines such as Speech2Spikes~\cite{stewart23_speech2spikes} , SIDC-KWS\cite{lim25_interspeech}, ED-sKWS~\cite{song24c_interspeech}, and approaching the performance of
delay-learning architectures such as DCLS-Delays~\cite{hammouamri24_iclr_delays}.
In the highly parameter-constrained configuration, the \textit{Tiny} model ($\sim$35k parameters) retains 89.8\% accuracy, matching or exceeding baselines that rely on substantially larger backbones.
These results highlight that improving the front end encoding can significantly reduce the parameter budget required to reach a given accuracy.   

\begin{table}[h]
  \centering
  \caption{Benchmarking against spiking KWS systems on GSC-v2 (35 classes).}
  \label{tab:kws_models}
  \small
  \setlength{\tabcolsep}{5pt}
  \begin{tabular}{@{}lcc@{}}
    \toprule
    Model & Params (K) & Acc. (\%) \\
    \midrule
    Ours (Large) & 1820 & 94.97 \\
    Ours (Small) & 699  & 92.64 \\
    Ours (Tiny)  & 35   & 89.80 \\
    \midrule
    DCLS-Delays \cite{hammouamri24_iclr_delays} & 2500 & 95.3 \\
    SIDC-KWS \cite{lim25_interspeech}           & 403  & 94.7 \\
    ED-sKWS \cite{song24c_interspeech}          & 307  & 93.1 \\
    SRNN+ALIF \cite{bittar22_frontiers_sg_baseline} & 222 & 92.5 \\
    Speech2Spikes \cite{stewart23_speech2spikes} & 410 & 89.5 \\
    \bottomrule
  \end{tabular}
\end{table}

\subsection{Interpretability: Separability vs.\ Reconstruction}
\label{ssec:analysis}

Does the encoder primarily preserve signal fidelity, or does it shape features for classification? To probe this, we freeze the trained encoder and train a classifier directly on its output spike representation. The learnable encoder yields 71.63\% test accuracy compared to 63.72\% for the fixed encoder, a gap of 8.24 percentage points, indicating improved class separability at the representation level. In addition, we analyze encoder training signals by inspecting per-band gradients and reconstruction residuals. The observed gradient patterns are not explained solely by reconstruction distortion. Specifically, The correlation between gradient magnitude and per-band spike activity is negligible ($r \approx -0.00018$), confirming that the encoder is not simply optimizing firing rate. In contrast, the correlation between gradient magnitude and reconstruction error is clearly positive ($r \approx 0.22$), indicating that regions where the encoder deviates more strongly from the original mel representation tend to receive larger training signals, suggesting that the learned quantization behavior is shaped by the downstream discriminative objective rather than signal fidelity alone.

\subsection{Sparsity and Energy Proxy}
\label{ssec:energy_results}
For the Large model, event activity is evaluated and energy is estimated using a hardware-agnostic compute proxy. We adopted the Synaptic operation count (SynOps) that reflects the number of synaptic operations effectively activated by spikes\cite{10.3389/fnins.2020.00662}. We emphasize that this provides a coarse estimate intended for relative comparison rather than a precise measurement of on-chip energy. \needspace{4\baselineskip} 
Activity is reported by collecting the network firing rate, and the event-driven operations are estimated as:
\begin{equation}
\mathrm{SynOps} = \mathrm{DenseOps} \times \mathrm{SpikingRate}.
\end{equation}

Here, \(\mathrm{DenseOps}\) denotes the nominal dense operation count under full activation, and \(\mathrm{SpikingRate}\) is the measured average firing rate. Accumulate operations (\emph{AC}) for SNN synaptic events are also computed. These operation counts are then converted into Joules using Horowitz-style 45\,nm energy estimates\cite{6757323}. As shown in Table~\ref{tab:compute_learning}, the Large model achieves 95.5\% global sparsity, reducing operations from 82.9\,M dense to 3.9\,M active SynOps, with an estimated active energy of 16\,\(\mu\)J.

\begin{table}[t]
\centering
\caption{Compute proxy on GSC-v2 for the Large model (45\,nm estimates).}
\label{tab:compute_learning}
\begin{tabular}{l c}
\toprule
\multicolumn{2}{c}{\textbf{Compute Proxy}} \\
\midrule
Global Sparsity (\%) & 95.5 \\
Dense Ops (Nominal) & 82.9 M \\
Active Ops (Event-driven) & 3.9 M \\
Estimated Active Energy & 16 $\mu$J \\
\bottomrule
\end{tabular}
\end{table}

\subsection{Benchmarking Learning Rules}
\label{ssec:learning_results}

Finally, we compare surrogate-gradient BPTT against DFA on the same architecture, learnable encoder and training conditions, building on prior demonstrations that DFA \cite{lillicrap16_natcomm_feedback_alignment, unknown} can support learning in spiking neural networks while avoiding symmetric weight transport.
This controlled comparison establishes a clear accuracy--hardware-efficiency trade-off between surrogate-gradient BPTT and DFA. While DFA simplifies credit assignment and avoids symmetric weight transport, it achieves lower accuracy (91.5\%) than surrogate-gradien BPTT (94.97\%) under identical conditions. This performance gap highlights both the promise and current limitations of local learning rules for neuromorphic speech applications and motivates further research to close the remaining accuracy gap without sacrificing hardware efficiency.

%==================================================================================
\section{Conclusion}
\label{sec:conclusion}

We introduce a learnable residual speech-to-spike encoder that replaces fixed Step-Forward thresholds with trainable coarse and fine step sizes and jointly optimized with an R-LIF backbone. The proposed approach improves classification accuracy while reducing input spike activity in neuromorphic keyword spotting. We further provided a controlled benchmark of learning rules, showing that Direct Feedback Alignment can train the same spiking keyword spotting model to high accuracy but remains below surrogate-gradient BPTT. This result highlights a fundamental trade-off between hardware-friendly local learning and optimal performance. 
Future work includes tightening hardware-aware energy estimates through more fine-grained, layer-level compute proxies, and exploring sparsity-constrained local learning rules that aim to close the DFA–BPTT accuracy gap without increasing spike activity.

\section{Acknowledgments}

Taharim Rahman Anon contributed to this work during her internship at PI LLC. The authors thank PI LLC (Sapporo, Hokkaido, Japan) for providing the GPU resources that supported the experiments in this study.

\section{Generative AI Use Disclosure}
In this paper, we have utilized ChatGPT (OpenAI: GPT-5.2) only to assist with minor  editing and polishing of the manuscript after the core scientific content and main ideas of the work had been developed and written by the authors. The specific assistance of AI tools includes editing and formatting equations into LaTeX, including grammar, spelling, and overall readability to ensure the textual consistency. No generative AI tool was used to generate the primary research contributions, technical ideas, experimental design, or results. All authors reviewed and take full responsibility for the  originality, accuracy, and integrity of the paper.
%==================================================================================
\bibliographystyle{IEEEtran}
\bibliography{mybib}

\end{document}